\pdfoutput=1
\documentclass[11pt]{article}

\usepackage{ACL2023}

\usepackage{times}
\usepackage{latexsym}

\usepackage{url}
\usepackage{soul}         % For highlighting text
\usepackage{geometry}
\usepackage{amsmath, amssymb}
\usepackage{graphicx}
\usepackage{hyperref}
\usepackage{enumitem}
\usepackage{booktabs}     % For better horizontal lines (\toprule, \midrule, \bottomrule)
\usepackage{multirow}     % For multirow cells if needed in the future
\usepackage{array}        % For customizing column alignment
\usepackage{caption}      % For table captions
\usepackage{natbib}       % For better bibliography handling
\usepackage{colortbl}       % Provides row and cell coloring

\usepackage[T1]{fontenc}

\usepackage[utf8]{inputenc}

\usepackage{microtype}

\usepackage{inconsolata}

\title{Towards Scalable and Cross-Lingual Specialist Language Models for Oncology}

\author{
  Morteza Rohanian \quad Tarun Mehra \quad Nicola Miglino \quad Farhad Nooralahzadeh \\
  \textbf{Michael Krauthammer} \quad \textbf{Andreas Wicki} \\
  University of Zurich and University Hospital Zurich, Switzerland  \\
  \texttt{morteza.rohanian@uzh.ch} \\ % Correspondence email only
}

\begin{document}
\maketitle
\begin{abstract}

Clinical oncology generates vast, unstructured data that often contain inconsistencies, missing information, and ambiguities, making it difficult to extract reliable insights for data-driven decision-making. General-purpose large language models (LLMs) struggle with these challenges due to their lack of domain-specific reasoning, including specialized clinical terminology, context-dependent interpretations, and multi-modal data integration.
We address these issues with an oncology-specialized, efficient, and adaptable NLP framework that combines instruction tuning, retrieval-augmented generation (RAG), and graph-based knowledge integration. Our lightweight models prove effective at oncology-specific tasks, such as named entity recognition (e.g., identifying cancer diagnoses), entity linking (e.g., linking entities to standardized ontologies), TNM staging, document classification (e.g., cancer subtype classification from pathology reports), and treatment response prediction.
Our framework emphasizes adaptability and resource efficiency. We include minimal German instructions, collected at the University Hospital Zurich (USZ), to test whether small amounts of non-English language data can effectively transfer knowledge across languages. This approach mirrors our motivation for lightweight models, which balance strong performance with reduced computational costs, making them suitable for resource-limited healthcare settings. We validated our models on oncology datasets, demonstrating strong results in named entity recognition, relation extraction, and document classification.

\end{abstract}

\section{Introduction}

Clinical oncology and related disciplines such as radiology or pathology often capture patient-related information in an unstructured or semi-structured way. At the same time, there is an increasing need to use real-world data to enable data-driven therapy decisions as a strategy that complements standardized evidence-based (study-informed) decision making. 

In the typical healthcare setting, oncologists must gather vast amounts of  information  from different data sources, including radiology images and reports, pathology reports, molecular analyses, clinical notes, and patient histories. They rely on these diverse sources to guide diagnosis, the assessment of prognosis and stage, and the decision on therapy. However, much of this data is in free text format within electronic health records (EHR) \cite{pardoll2012blockade, topol2019high}. Clinicians waste time and resources as they parse these notes by hand. This leads to slow, inconsistent, and error-prone decision-making, especially in resource-limited environments \cite{bedogni2009clinical}.

Natural language processing (NLP) offers tools to extract insights from free-text clinical records. Rule-based systems and machine learning methods with hand-engineered features have successfully identified entities such as diseases and treatments \cite{alawad2020automatic}. These methods fail to handle the nuanced language and variability of clinical data. Pretrained language models (LMs), such as BERT \cite{devlin2018bert}, BioBERT \cite{lee2020biobert}, and ClinicalBERT \cite{alsentzer2019publicly}, improve performance on tasks like entity recognition and literature mining by leveraging large biomedical corpora \cite{gu2021domain, huang2020tabtransformer, rohanian2023effectiveness, rohanian2024lightweight}. Despite these advances, such models focus primarily on classification, lack flexible reasoning capabilities, and are limited in their ability to generate coherent text for summarization or prediction \cite{ruder2017overview}. Moreover, these models predominantly support English, overlooking the multilingual requirements of many healthcare systems.

Large language models (LLMs), such as GPTs \cite{brown2020language} and LLaMA \cite{touvron2023llama}, overcome some of these limitations by handling diverse tasks and adapting to new domains with minimal labeled data. Researchers have used them to summarize medical records, answer questions, and support clinical decisions \cite{singhal2023large, saab2024capabilities}. General-purpose LLMs often fail in specialized fields like oncology. They lack domain-specific knowledge, produce inconsistent reasoning \cite{wu2024medical, hu2024grag}, and require substantial computational resources, which many healthcare institutions cannot afford. Lightweight models provide a practical alternative by delivering strong performance with significantly reduced resource requirements.

Recent research has adapted LLMs for oncology-specific applications, often addressing single tasks such as named entity recognition (NER) or relation extraction \cite{alawad2021integration, zhou2022cancerbert, nishiozero, fujimotoclassification}. However, these approaches lack scalability and multilingual flexibility. Newer methods integrate biomedical corpora, retrieval mechanisms, and parameter-efficient fine-tuning to handle complex tasks. Some studies have curated large corpora (e.g., from the TCGA dataset) to build prognostic models or classify cancer subtypes, but these often rely on manual feature engineering or rule-based systems \cite{alawad2021integration}. Other works used transformer-based models for TNM extraction, disease coding, or limited classification tasks \cite{kefeli2024generalizable}.

We propose an oncology-specialized NLP framework that combines lightweight models, bilingual adaptability, and advanced reasoning techniques. Given the Swiss healthcare system’s nature, incorporating German alongside English ensures the framework can address the linguistic diversity encountered in clinical practice at institutions like USZ. We curated minimal German instructions from clinical queries at the University Hospital Zurich (USZ) and systematically varied their number (100, 200, 400) to test whether small amounts of bilingual data can transfer domain-specific knowledge effectively across languages. Both bilingual adaptability and lightweight models align with our overarching goal of creating scalable NLP systems that can adapt to diverse healthcare environments, from large hospitals to resource-limited clinics.

\noindent Our framework tries to solve key challenges in oncology NLP by integrating instruction tuning, retrieval-augmented generation, and graph-based reasoning. Each component targets specific issues in processing clinical data.

Instruction tuning improves the accuracy of lightweight models for oncology-specific tasks. Models handle named entity recognition, relation extraction, TNM staging, and treatment response prediction with precision. Bilingual instructions in English and German align with real clinical use cases such as ICD-10 coding and treatment classification. Testing this variation reveals how small bilingual datasets transfer knowledge across languages and strengthen cross-lingual adaptability.

RAG improves outputs by retrieving relevant clinical data from trusted sources. External datasets such as MIMIC-IV and curated German oncology reports add real-time context to the model’s responses. The retrieval process connects queries with factual information from oncology corpora. Using hierarchical methods, RAG retrieves critical details efficiently without overwhelming the input with unnecessary context.

Graph-based reasoning ensures outputs are reliable and factually grounded. A knowledge graph integrates resources like UMLS, linking extracted entities to verified medical facts. Relationships between entities, such as treatments and stages, are organized as nodes and edges. Triple graph construction connects entities to authoritative references, reducing ambiguity and improving reasoning. This process strengthens the clinical reliability of model-generated outputs.

Lightweight LLaMA variants (LLaMA-2-7B, LLaMA-3.1-8B, LLaMA-3.2-1B, and LLaMA-3.2-3B) combine these methods to balance efficiency and performance. The framework adapts to resource-limited clinical environments while maintaining high accuracy and flexibility across oncology-specific applications.

Our contributions are as follows:  

\begin{enumerate}[noitemsep,topsep=2pt,left=0pt]  
    \item \textbf{Oncology-Specialized Modeling:} Lightweight models fine-tuned for oncology tasks like TNM staging, named entity recognition, relation extraction, document classification, and treatment prediction. Benchmarks include datasets like NCBI-Disease, i2b2-2010, and labeled subsets of TCGA.  
    \item \textbf{Multilingual Adaptability:} Minimal German instructions collected from USZ improve cross-lingual performance on ICD-10 coding and TNM staging. The bilingual framework supports diverse healthcare systems by addressing multilingual requirements.
    \item \textbf{Model Efficiency:}  Lightweight models such as LLaMA-2-7B deliver high accuracy with lower computational costs. This ensures advanced NLP tools remain accessible to institutions with limited resources.
    \item \textbf{Task Adaptability:}  The framework applies to diverse tasks, including relation extraction, document classification, and multilingual ICD-10 coding. Models adapt to new domains and tasks. 
\end{enumerate}

The integration of instruction tuning, RAG, and graph-based reasoning provides oncology NLP systems that deliver accurate, efficient, and context-aware solutions for multilingual and resource-limited settings.

\begin{figure*}[t]
    \centering
    \includegraphics[width=\textwidth]{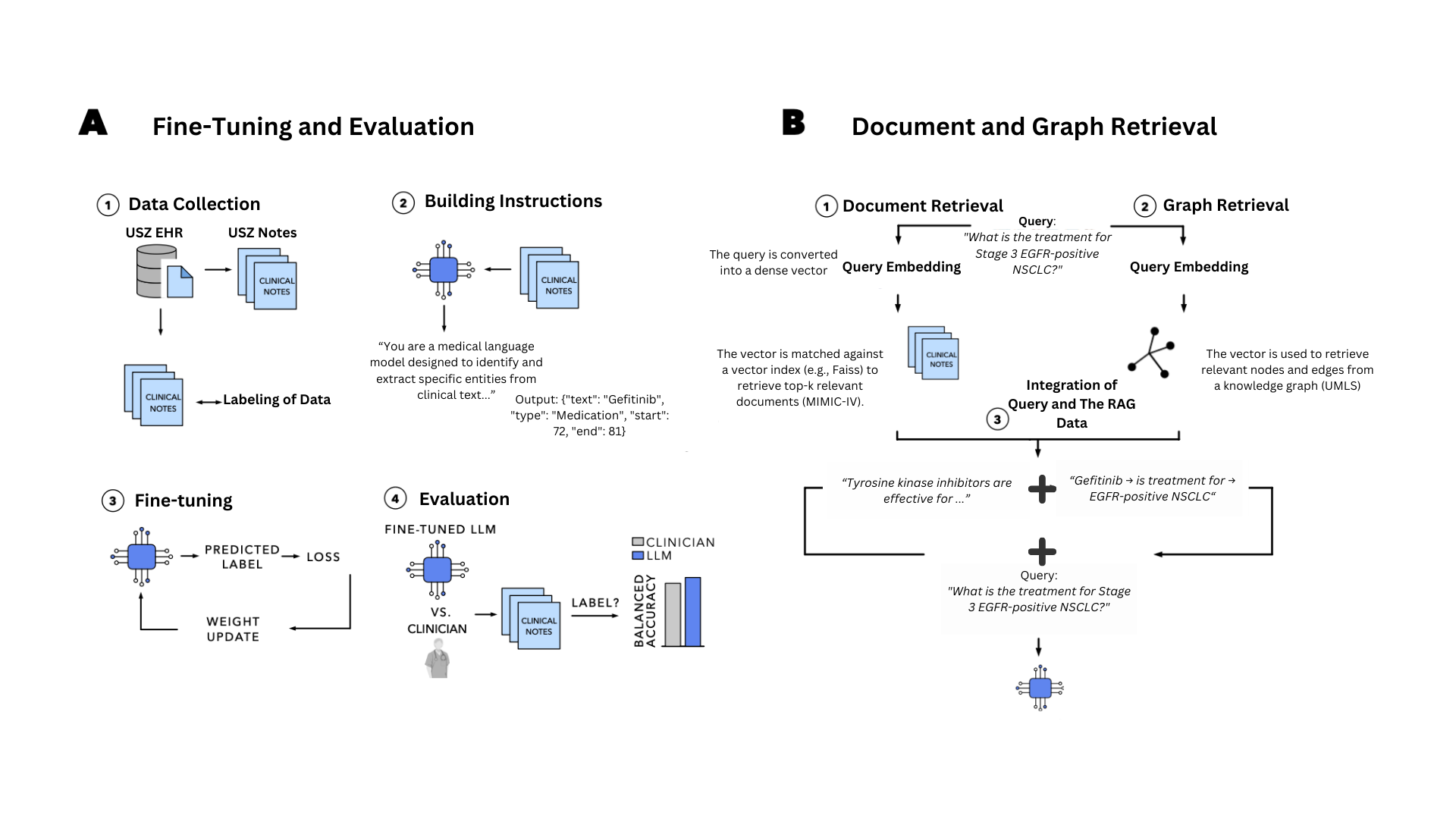}
    \caption{(A) Fine-tuning and evaluation workflow: This panel shows the process of data collection, instruction building, fine-tuning, and evaluation against clinician annotations. (B) Document and graph retrieval: This panel highlights the integration of document retrieval and graph-based reasoning for query-based inference.}
    \label{fig:fine_tuning_retrieval}
\end{figure*}

\begin{table*}[t!]
\caption{Instruction Tuning Examples for Oncology Tasks}
\label{tab:instruction_tuning_examples}
\centering
\renewcommand{\arraystretch}{1.4}
\setlength{\tabcolsep}{5pt} % Adjust column spacing
\resizebox{\textwidth}{!}{%
\begin{tabular}{|>{\columncolor[HTML]{D9EAF7}}p{3cm}|>{\columncolor[HTML]{F9F9F9}}p{5cm}|p{6cm}|>{\columncolor[HTML]{E2F0CB}}p{3cm}|}
\hline
\textbf{Task} & \textbf{Instruction} & \textbf{Input Text} & \textbf{Output} \\ \hline

\textbf{Hallmarks of Cancer (HoC)} &  
As a medical expert, assess the clinical text for cancer hallmarks. Assign one or more labels from the list: Sustaining proliferative signaling (PS), Enabling replicative immortality (RI), Inducing angiogenesis (A), Genome instability \& mutation (GI), Tumor-promoting inflammation (TPI), ... &  
\textit{Taken together, the present study clearly shows the synergistic anti-inflammatory as well as anti-oxidative stress effects of CUR and PUFA.} &  
Tumor-promoting inflammation (TPI) \\ \hline

\rowcolor[HTML]{FCF3CF}
\textbf{Natural Language Inference (MedNLI)} &  
Evaluate the connection between two clinical sentences and classify them into one of these categories: Contradiction (if the sentences conflict), Neutral (if no logical association), or Entailment (if one sentence logically implies the other)... &  
\textbf{Sentence 1:} \textit{Lung cancer as above s/p pneumonectomy} \newline \textbf{Sentence 2:} \textit{History of smoking.} &  
Neutral \\ \hline

\textbf{Relationship Extraction (i2b2-2010)} &  
In the clinical text, your objective is to identify relationships between medical problems, treatments, and tests. Medical problems are tagged as @problem\$, medical tests as @test\$, and treatments as @treatment\$. Classify the relationship as: Treatment is administered for medical problem (TrAP)... &  
\textit{His past medical history is significant for prostate cancer, benign prostatic hypertrophy, hypothyroidism, status post @treatment\$ for @problem\$, chronic painless hematuria, degenerative joint disease, and history of a murmur.} &  
TrAP \\ \hline

\rowcolor[HTML]{E8F8E0}
\textbf{Named Entity Recognition (NER)} &  
Your mission is to tag disease-related Named Entities in the text using the BIO labeling scheme. When you encounter a disease-related phrase, mark the start with B (Begin) and continue with I (Inner) ... &  
\textit{Its role in the therapy of glomerulonephritis, autoimmunity, cystic renal diseases and renal cancer is under investigation.} &  
\texttt{... cystic: B, renal: I, diseases: I, and: O, renal: B, cancer: I...} \\ \hline

\end{tabular}%
}
\end{table*}

\section{Data Sources}

We use a combination of bilingual clinical datasets and diverse public benchmarks to fine-tune and evaluate our oncology NLP framework. These datasets enable the exploration of bilingual adaptability, cross-lingual generalization, and task scalability.

\subsection{DUP-127 Clinical Dataset}

We include the DUP-127 dataset, a German oncology dataset containing structured and unstructured clinical data. This dataset aligns with our goal of creating a bilingual NLP framework to address the linguistic diversity in clinical practice, particularly in Swiss healthcare. The dataset was established within the framework of the Swiss Personalized Health Network (SPHN), an initiative supported by the Swiss State Secretariat for Education, Research and Innovation (SERI). It encompasses around 110 distinct structured datapoints such as ICD diagnoses, TNM annotations, and medications extracted from the electronic health care records of patients with cancer. Data were collected using the SPHN reference dataset (version 2021.1) . The dataset was represented with a Resource Description Framework (RDF) schema, encrypted, and securely transferred to the data repositories of the participating universities (BioMedIT network). To ensure semantic harmonization, a set of semantic rules written in shapes constraint language (SHACL) was defined and distributed together with the ontology. To prevent erroneous content and ensure interoperability, we integrated a set of queries in a data validation pipeline. Integrity checks were carried out to investigate possible missing data points and inconsistencies in patient timelines. Metastatic treatment lines were reconstituted according to progression dates. Drugs were grouped into regimens according to their administration dates and manually corrected by medical experts 
Unstructured elements, such as treatment histories, radiology reports as well as histology reports and genomic profiles, were manually annotated by expert physicians during initial data preparation. Diagnoses and genomic information link directly to corresponding free-text records using patient IDs, ensuring integration between structured and unstructured data. The SPO protocol (No. 2020-00347) was approved by the Northwest and Central Swiss Ethics Committee (EKNZ) and ratified by the local ethics committees (CCER, CER-VD, Kantonale Ethikkommission Bern, Kantonale Ethikkommission Zürich). 

\subsection{Public Datasets}

We fine-tune and evaluate our model on a range of tasks that capture the complexity of oncology practice. This multi-dataset approach reflects the adaptability of our framework across oncology-specific tasks and supports its scalability to different clinical challenges.

\textbf{NER:} We use NCBI-Disease \cite{dougan2014ncbi}, BC5CDR (Disease/Chem) \cite{li2016biocreative}, BC2GM \cite{ando2007biocreative}, JNLPBA \cite{collier2004introduction}, and i2b2-2012 \cite{uzuner20112010} to test how well the model extracts biomedical entities such as diseases, chemicals, or genes from text. These datasets focus on biomedical literature and primarily employ the standard BIO (Beginning-Inside-Outside) labeling scheme.  

\textbf{Relation Extraction:} i2b2-2010 \cite{uzuner20112010} and GAD \cite{bravo2015extraction} measure how well the model links genes, diseases, and treatments. This step tests the model’s ability to identify relations, for example, a gene-disease association or a drug-disease treatment link. The i2b2-2010 dataset centers on clinical narratives, where relationships are defined between problems, test results, and treatments. 

\textbf{NLI:} MedNLI \cite{romanov2018lessons} tests logical reasoning about clinical statements, requiring the model to determine whether a conclusion follows logically from given premises. This task is particularly relevant in oncology, where clinicians must reconcile conflicting findings from reports, pathology notes, or imaging summaries. For instance, determining whether a pathology report implies disease progression based on an imaging report involves reasoning over subtle textual cues. 

\textbf{Document Classification:} Document classification addresses the task of assigning labels to entire texts, such as clinical reports, based on their content. We use the \textit{Hallmarks of Cancer (HoC)} \cite{baker2016automatic} dataset and the \textit{TCGA Pathology Report Dataset} \cite{kefeli2024generalizable}cite for these experiments.

The Hallmarks of Cancer dataset provides multi-class labels aligned with ten canonical hallmarks of cancer, including sustained proliferative signaling, immune evasion, and genomic instability. These categories represent critical biological processes that drive cancer progression. By applying these labels, the model learns to classify biomedical literature according to underlying cancer-related themes.

The TCGA Pathology Report dataset grounds this classification in clinical practice. It includes 9,523 pathology reports spanning 32 distinct cancer types, each processed through OCR and careful post-processing. Beyond cancer-type classification, the TCGA reports include TNM staging annotations (T1–T4, N0–N3, M0–M1). TNM staging provides essential prognostic information and guides treatment decisions. We split this dataset into 70\% training, 15\% validation, and 15\% test, ensuring a balanced approach to model development and performance evaluation.

We also incorporate the \textit{MSK-IMPACT} \cite{zehir2017mutational} dataset, a curated resource from Memorial Sloan Kettering Cancer Center. It includes 1,479 patients treated with systemic immune checkpoint blockade (ICB). This dataset provides binary labels for treatment response, where patients are categorized as responders or non-responders based on clinical response criteria, such as the RECIST v1.1 guidelines. Responders include both complete responders (CR), defined as the disappearance of all target lesions, and partial responders (PR), defined as at least a 30\% decrease in the sum of the diameters of target lesions. Non-responders encompass patients with stable disease (SD) or progressive disease (PD).

\section{Methodology}

Our methodology transforms pretrained language models into specialized oncology tools by integrating instruction tuning, retrieval-augmented generation (RAG), and graph-based knowledge integration. 
In Figure~\ref{fig:fine_tuning_retrieval}, we illustrate the fine-tuning process (Panel A) and the document and graph retrieval mechanisms (Panel B). Panel A demonstrates the end-to-end workflow for building labeled datasets, constructing instructions, and fine-tuning lightweight models. Panel B highlights how the system integrates document retrieval, graph-based reasoning, and query embeddings to generate clinically relevant responses. Together, these steps form the core of our methodology for transforming general-purpose LLMs into oncology-specialized tools.

These components enable the models to process complex oncology data, reason about medical facts, and generate precise predictions for clinical workflows. By emphasizing bilingual adaptability through minimal German instructions and resource-efficient lightweight models, we ensure our approach scales across multilingual and resource-limited healthcare environments.

\subsection{Instruction Tuning Across Languages}

To fine-tune our lightweight generative language models (LLaMA-2-7B, LLaMA-3.1-8B, LLaMA-3.2-1B, and LLaMA-3.2-3B), we use curated instruction-response pairs in English and German. These instructions simulate real-world oncology queries, such as identifying cancer-related entities, TNM staging annotations, or extracting treatment protocols. Each instruction-response pair provides structured outputs, such as JSON-formatted annotations specifying entity types, attributes, and their spans within the text. For instance, a tumor-related entity recognition query might yield outputs categorizing “lung cancer” or “EGFR-positive adenocarcinoma” with attributes like diagnosis date or molecular markers. Table~\ref{tab:instruction_tuning_examples} provides examples of instructions used across different oncology tasks, highlighting their diversity and task-specific objectives. These examples demonstrate how instructions align with tasks like named entity recognition, natural language inference, and relation extraction, ensuring task relevance and improving model generalization \cite{rohanian2024exploring}.

To evaluate cross-lingual adaptability, we augment public datasets with minimal German instructions, ranging from 100 to 400 examples. These instructions cover tasks such as ICD coding, TNM staging, and treatment annotation. Training minimizes the instruction tuning loss:
\[
\mathcal{L}_{\text{tuning}} = -\frac{1}{N} \sum_{i=1}^N \log P_\theta(y_i \mid x_i, \text{instruction}),
\]
where \(x_i\) represents the input text, “instruction” specifies the task, and \(y_i\) is the expected response. Cross-validation splits are applied to ensure generalization to unseen instructions and languages.

\subsection{Retrieval-Augmented Generation (RAG)}

Oncology workflows often require reasoning over large, diverse, and evolving datasets. To address this complexity, we integrate retrieval-augmented generation (RAG), which grounds model responses in external knowledge.  We use a sentence embedding model, fine-tuned for oncology-specific tasks, to encode user queries (\(Q\)) and candidate documents \(D\) into dense vector representations. These embeddings capture semantic similarity between clinical terms and contexts. To store and index these embeddings efficiently, we use the FAISS (Facebook AI Similarity Search) library \cite{johnson2019billion}. FAISS provides high-speed similarity searches across large document collections, enabling real-time retrieval and processing of oncology data. User queries \(Q\) and candidate documents \(D\) are encoded into dense vector representations, with cosine similarity determining their relevance:
\[
\text{sim}(Q, D) = \frac{\phi(Q) \cdot \phi(D)}{\|\phi(Q)\| \|\phi(D)\|}.
\]
The top-k most relevant documents, selected based on similarity scores, are appended to the model’s input. These documents are drawn from external datasets MIMIC-IV and curated German oncology discharge reports, ensuring that model responses remain accurate, evidence-based, and context-aware.

We incorporate semantic document chunking to improve retrieval efficiency. Oncology documents, often lengthy and complex, are segmented into smaller, contextually coherent chunks. We encode each chunk using a sentence embedding model fine-tuned for oncology-specific tasks. The resulting dense vector representations are indexed in the FAISS, enabling fast and scalable similarity-based searches. This hybrid approach uses paragraph-based splitting combined with semantic similarity analysis, ensuring that each chunk retains topical coherence. By storing these chunks independently in the FAISS index, the system ensures that even detailed oncology data is processed and retrieved with high granularity and contextual relevance. Semantic chunking also aligns with graph-based knowledge integration by mapping extracted entities to corresponding graph nodes.

We optimize retrieval further using a hierarchical U-Retrieval strategy. High-level clinical tags, such as tumor stage, disease type, or treatment categories, guide the initial retrieval, reducing the document pool to a manageable size. The system then iteratively integrates broader contextual summaries, balancing precision with global context awareness. This multi-layered retrieval enables comprehensive reasoning over complex oncology-specific scenarios.

\begin{table*}[t!]
\caption{Performance of Models Across Biomedical Tasks with Different Configurations}
\label{tab:model_performance}
\centering
\resizebox{\textwidth}{!}{
\begin{tabular}{llcccccccc}
\toprule
\multicolumn{2}{c}{\textbf{Model Configuration}} & \textbf{NCBI-Disease} & \textbf{BC5CDR-Disease} & \textbf{BC5CDR-Chem} & \textbf{BC2GM} & \textbf{JNLPBA} & \textbf{i2b2-2012} & \textbf{i2b2-2010} & \textbf{MedNLI} \\
\cmidrule(r){1-2} \cmidrule(l){3-10}
\textbf{Type} & \textbf{Model} & NER & NER & NER & NER & NER & NER & RE & NLI \\
\midrule
\multicolumn{10}{l}{\textbf{Base LLM}} \\
 & LLaMA-2-7B & 85.69 & 83.12 & 93.77 & 77.40 & 79.67 & 79.66 & 90.01 & 88.76 \\
 & LLaMA-3.1-8B & 86.33 & 83.86 & 93.45 & 79.95 & 79.78 & 80.58 & 90.83 & 88.09 \\
 & LLaMA-3.2-1B & 85.58 & 82.48 & 92.41 & 77.30 & 79.63 & 79.67 & 89.59 & 86.63 \\
 & LLaMA-3.2-3B & 83.56 & 82.89 & 92.27 & 78.97 & 79.12 & 79.97 & 89.84 & 86.63 \\
\midrule
\multicolumn{10}{l}{\textbf{Instruction-Tuned}} \\
 & LLaMA-2-7B & 88.37 & 86.48 & 94.12 & 82.70 & 82.77 & 81.62 & 93.70 & 90.57 \\
 & LLaMA-3.1-8B & \textbf{89.50} & \textbf{87.64} & \textbf{94.82} & 84.41 & \textbf{83.60} & \textbf{81.92} & 93.26 & 90.56 \\
 & LLaMA-3.2-1B & 85.70 & 86.08 & 93.08 & 81.34 & 81.96 & 80.83 & 92.08 & 90.61 \\
 & LLaMA-3.2-3B & 85.43 & 86.08 & 93.27 & 81.70 & 81.99 & 80.88 & 92.57 & 89.88 \\
\midrule
\multicolumn{10}{l}{\textbf{+RAG}} \\
 & LLaMA-2-7B & 88.17 & 86.61 & 94.38 & 82.76 & 82.54 & \textbf{81.92} & 92.22 & \textbf{91.89} \\
 & LLaMA-3.1-8B & 88.85 & 87.50 & 94.77 & 84.71 & 83.01 & 81.21 & 92.95 & 91.08 \\
 & LLaMA-3.2-1B & 85.74 & 86.45 & 93.29 & 82.27 & 81.24 & 80.39 & 91.82 & 90.20 \\
 & LLaMA-3.2-3B & 85.47 & 86.03 & 93.18 & 82.37 & 81.90 & 80.77 & 91.07 & 90.61 \\
\midrule
\multicolumn{10}{l}{\textbf{+Graph-RAG}} \\
 & LLaMA-2-7B & 88.26 & 86.42 & 94.67 & 84.06 & 82.29 & 81.80 & 93.63 & 91.19 \\
 & LLaMA-3.1-8B & 88.79 & 87.32 & 94.40 & \textbf{84.84} & 83.56 & \textbf{81.92} & 93.50 & 91.87 \\
 & LLaMA-3.2-1B & 87.53 & 86.49 & 93.22 & 82.64 & 81.41 & 80.39 & 92.62 & 90.72 \\
 & LLaMA-3.2-3B & 87.37 & 86.53 & 93.90 & 83.59 & 82.09 & 80.26 & 92.57 & 90.58 \\
\bottomrule
\end{tabular}
}
\end{table*}

\subsection{Graph-Based Knowledge Integration}

To enhance factual reliability and interpretability, we integrate a domain-specific knowledge graph \(G\), constructed from standardized resources UMLS, SNOMED-CT, and ICD-10. This graph encodes entities as nodes and their relationships as edges:
\[
G = \{(v_i, e_{ij}, v_j) \mid v_i, v_j \in V, \; e_{ij} \in E\},
\]
where \(v_i\) and \(v_j\) represent medical entities (e.g., “adenocarcinoma” or “Osimertinib”), and \(e_{ij}\) represents relationships (e.g., “treated\_with”).

Graph enrichment occurs through triple graph construction, linking retrieved entities to authoritative references and professional definitions:
\[
\text{Triple} = [\text{entity}, \text{source}, \text{definition}].
\]
For instance, a TNM stage extracted from text is mapped to corresponding UMLS nodes and linked to oncology treatment guidelines, ensuring outputs remain grounded in verified medical knowledge.

To encode the graph, we employ a two-step process:
\begin{enumerate}
    \item \textbf{Node Encoding:} Each node is represented as a dense vector embedding using a pretrained graph embedding model TransE \cite{bordes2013translating}. These embeddings capture the semantic meaning of entities based on their attributes and the structure of the graph. For example, the embedding for “adenocarcinoma” encodes its connections to treatments, symptoms, and associated genes.
    \item \textbf{Edge Encoding:} Relationships (edges) between nodes are represented as directional vectors. These are computed by applying transformation functions to the embeddings of the connected nodes. For instance, the edge “treated\_with” between a disease node and a medication node reflects the nature and direction of the relationship.
\end{enumerate}

Hierarchical tagging further improves graph efficiency and interpretability. Each graph node is tagged with categories such as “Symptoms,” “Medications,” or “Patient History,” creating a multi-level abstraction. During inference, the model accesses relevant graph layers, ensuring fast and precise retrieval for tasks that require high-level summaries and fine-grained details.

The combined encoding of nodes and edges enables efficient traversal and reasoning over the graph. By embedding the graph in a high-dimensional space, the model can retrieve semantically similar nodes and relations, supporting robust and context-aware clinical predictions.

\subsection{Model Implementation and Evaluation Metrics}

The instruction tuning, RAG, and graph-based reasoning components are integrated into lightweight LLaMA variants, creating a unified inference pipeline. Scalability is evaluated by varying the number of German instructions and the model size. Minimal German instructions (100–400 examples) are used to test cross-lingual adaptability, highlighting how small bilingual datasets influence performance. During training, we systematically vary the instructions to improve the model’s adaptability. Lightweight models are compared with larger variants to assess their performance in resource-constrained environments. 

We evaluate the framework’s performance using metrics tailored to specific tasks. For entity recognition, relation extraction, and document classification, we report the F1 score. For imbalanced datasets like TCGA-C, we use the area under the precision-recall curve (AU-PRC) to emphasize performance in uneven class distributions. Binary tasks, such as TNM staging and treatment response prediction, are evaluated using the area under the curve (AUC).

\begin{table*}[t!]
\caption{Performance of Models on English and Multilingual Tasks}
\label{tab:english_multilingual_tasks}
\centering
\resizebox{\textwidth}{!}{
\begin{tabular}{llccccccccccc}
\toprule
\multicolumn{2}{c}{\textbf{Model Configuration}} & \textbf{HoC} & \textbf{TCGA-C} & \textbf{TCGA-T} & \textbf{TCGA-N} & \textbf{TCGA-M} & \textbf{MSK-IMPACT} & \textbf{ICD-10} & \textbf{DUP-T} & \textbf{DUP-N} & \textbf{DUP-M} & \textbf{SNOMED} \\
\cmidrule(r){1-2} \cmidrule(l){3-13}
\textbf{Type} & \textbf{Model} & EN & EN & EN & EN & EN & EN & DE & DE & DE & DE & DE \\
\midrule
\multicolumn{13}{l}{\textbf{Base LLM}} \\
 & LLaMA-2-7B & 79.32 & 0.89 & 0.92 & 0.91 & 0.73 & 0.78 & 77.02 & 0.81 & 0.78 & 0.73 & 0.78 \\
 & LLaMA-3.1-8B & 80.12 & 0.89 & 0.92 & \textbf{0.92} & 0.73 & 0.78 & 78.58 & 0.81 & 0.78 & \textbf{0.74} & 0.78 \\
 & LLaMA-3.2-1B & 80.01 & 0.88 & 0.91 & 0.90 & 0.71 & 0.77 & 76.16 & 0.77 & 0.75 & 0.71 & 0.70 \\
 & LLaMA-3.2-3B & 80.09 & 0.88 & 0.91 & 0.90 & 0.71 & 0.77 & 76.53 & 0.75 & 0.77 & 0.71 & 0.72 \\
\midrule
\multicolumn{13}{l}{\textbf{Instruction-Tuned}} \\
 & LLaMA-2-7B & 82.34 & 0.89 & 0.92 & 0.92 & 0.74 & \textbf{0.79} & 82.84 & 0.82 & 0.80 & 0.73 & 0.82 \\
 & LLaMA-3.1-8B & 83.12 & \textbf{0.90} & 0.93 & 0.92 & 0.74 & 0.78 & 83.45 & 0.83 & 0.80 & 0.75 & 0.81 \\
 & LLaMA-3.2-1B & 82.03 & 0.88 & 0.91 & 0.90 & 0.73 & 0.77 & 80.04 & 0.78 & 0.77 & 0.72 & 0.72 \\
 & LLaMA-3.2-3B & 82.11 & 0.88 & 0.91 & 0.90 & 0.73 & 0.77 & 80.08 & 0.75 & 0.77 & 0.73 & 0.72 \\
\midrule
\multicolumn{13}{l}{\textbf{+RAG}} \\
 & LLaMA-2-7B & 82.32 & 0.89 & 0.93 & \textbf{0.93} & 0.74 & \textbf{0.80} & 82.07 & 0.83 & \textbf{0.81} & 0.75 & 80.00 \\
 & LLaMA-3.1-8B & 83.12 & 0.89 & \textbf{0.94} & \textbf{0.93} & \textbf{0.75} & \textbf{0.80} & 82.17 & \textbf{0.83} & 0.80 & 0.74 & 81.00 \\
 & LLaMA-3.2-1B & 82.12 & 0.88 & 0.92 & 0.91 & 0.73 & 0.78 & 79.87 & 0.80 & 0.77 & 0.72 & 0.72 \\
 & LLaMA-3.2-3B & 82.71 & 0.89 & 0.91 & 0.91 & 0.73 & 0.77 & 80.05 & 0.80 & 0.77 & 0.73 & 0.72 \\
\midrule
\multicolumn{13}{l}{\textbf{+Graph-RAG}} \\
 & LLaMA-2-7B & 83.11 & 0.91 & 0.94 & \textbf{0.93} & \textbf{0.75} & 0.79 & 84.86 & 0.82 & 0.81 & 0.74 & 0.84 \\
 & LLaMA-3.1-8B & \textbf{83.84} & \textbf{0.91} & \textbf{0.94} & \textbf{0.93} & \textbf{0.75} & \textbf{0.80} & \textbf{85.75} & 0.82 & 0.80 & 0.74 & \textbf{0.85} \\
 & LLaMA-3.2-1B & 82.76 & 0.90 & 0.92 & 0.92 & 0.73 & 0.77 & 83.00 & 0.80 & 0.77 & 0.72 & 0.77 \\
 & LLaMA-3.2-3B & 83.04 & 0.90 & 0.92 & 0.91 & 0.73 & 0.78 & 84.18 & 0.80 & 0.78 & 0.73 & 0.79 \\
\bottomrule
\end{tabular}
}
\end{table*}

\section{Results}

We evaluated our instruction-tuned LLMs across biomedical and oncology tasks. Each model variant (\texttt{LLaMA-2-7B}, \texttt{LLaMA-3.1-8B}, \texttt{LLaMA-3.2-1B}, and \texttt{LLaMA-3.2-3B}) underwent a progression: base configuration, instruction tuning, RAG integration, and Graph-RAG enhancement. We observed general performance boosts at every stage especially with the oncology tasks. Larger models like \texttt{LLaMA-3.1-8B} achieved higher accuracy, but smaller models like \texttt{LLaMA-3.1-1B} remained competitive as required fewer resources.

Instruction tuning substantially increased F1 scores on standard NER benchmarks. For example, \texttt{LLaMA-3.1-8B} improved from 86.33\% to 89.50\% on \textit{NCBI-Disease} after tuning, while \texttt{LLaMA-2-7B} jumped from 83.12\% to 86.48\% on \textit{BC5CDR-Disease}. This tuning step aligned the models with domain-specific tasks and helped them recognize disease names, biomarkers, and chemical entities. Adding retrieval further refined results. On \textit{BC5CDR-Chem}, contextual information reduced confusion about similar chemical mentions. Graph-RAG then linked terms to the ontologies, which resolved ambiguities and improved NER performance on more complex datasets like \textit{JNLPBA}.

Instruction tuning also helped relation extraction. The models learned to link diseases with treatments or genetic variants. \texttt{LLaMA-2-7B} reached 93.70\% on \textit{i2b2-2010}, while \texttt{LLaMA-3.1-8B} achieved similar results. Using RAG and Graph-RAG, the models matched gene-disease pairs more precisely.

Natural language inference tasks like \textit{MedNLI} tested logical reasoning. Instruction-tuned \texttt{LLaMA-2-7B} improved from 88.76\% to 90.57\%. 

On oncology-specific tasks, the graph-based models excelled. For example, in TNM staging, Graph-RAG improved entity linking by referencing established oncology guidelines, boosting F1 scores by 2.5\%. This structured reasoning allowed the models to generate consistent, verifiable outputs even in complex staging scenarios. \texttt{LLaMA-3.1-8B} classified biomedical literature into canonical cancer hallmarks with an F1 of 83.84\%. Linking TNM attributes to known ontologies supported the model in assigning the correct category, raising F1 scores on T, N, and M labels. \texttt{LLaMA-2-7B}, even though smaller, benefited from Graph-RAG and produced high AUC values. 

\begin{figure*}[t!]
    \centering
    \includegraphics[width=\textwidth]{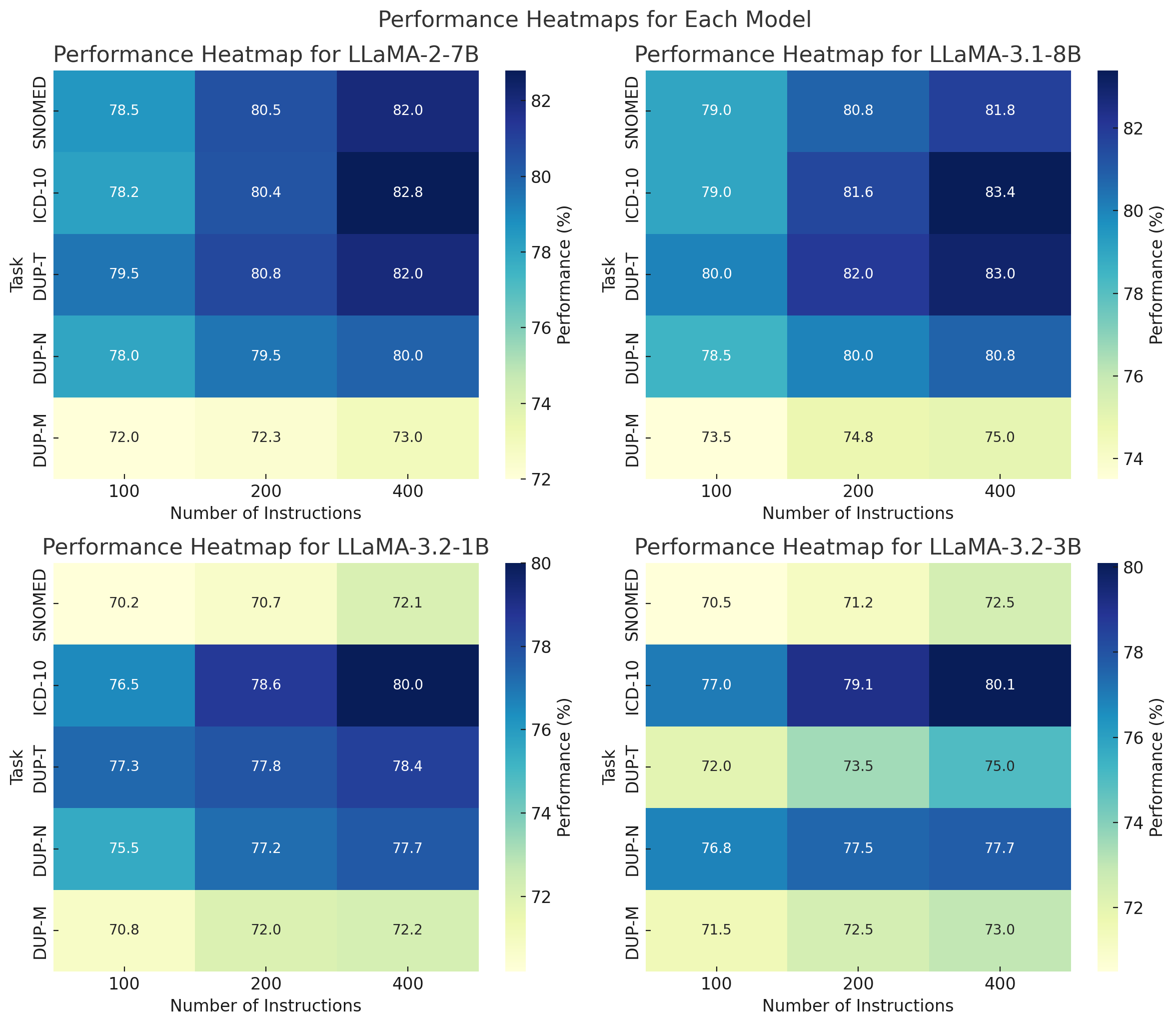}
    \caption{Performance scores for Instruction-Tuned Models with 100, 200, and 400 German instructions.}
    \label{fig:performance-heatmap}
\end{figure*}

Cross-lingual tests revealed the value of even minimal German instructions on tasks with the DUP-127 dataset (Figure~\ref{fig:performance-heatmap}). \texttt{LLaMA-2-7B} improved from 77.02\% to 82.84\% on \textit{ICD-10} coding by learning from a few hundred German instructions. The multilingual tuning also helped the \textit{SNOMED} classification and TNM staging in German. Figure~\ref{fig:performance-heatmap} highlights that gains peaked with 200 instructions, illustrating that even small bilingual datasets can enable cross-lingual generalization.

Scaling the number of German instructions from 100 to 400 led to incremental gains. Performance improvements peaked around 200 instructions, but complex tasks still benefited from 400. \texttt{LLaMA-3.1-8B}, with its larger capacity, took better advantage of these extra instructions. Smaller models also gained but reached a plateau sooner. \texttt{LLaMA-2-7B} maintained a balance between efficiency and accuracy, making it attractive for clinical environments with limited computational resources.

Overall, instruction tuning, retrieval, and graph-based reasoning worked together to produce a flexible model family. The models adapted to new tasks, integrated domain knowledge at inference time, and reasoned about complex medical concepts. They achieved these gains without overhauling internal parameters whenever new data emerged and instead leaned on retrieval and graphs to fetch needed facts on the fly.

\section{Discussion}

Our findings show the promise of combining instruction tuning, retrieval augmentation, and graph-based knowledge integration for oncology NLP. The incorporation of a few instructions in another language demonstrated the potential of cross-lingual capabilities. By using minimal bilingual training data, our approach bypassed the usual costs associated with large-scale multilingual training, offering a practical and scalable solution for global healthcare systems with diverse linguistic needs.

Retrieval augmentation added critical agility to the system, allowing the model to dynamically access up-to-date information at inference time instead of relying solely on static, parameter-encoded knowledge. This design enables models to adapt to evolving oncology guidelines and clinical practices, which often change multiple times a year. For example, retrieval mechanisms can help the model navigate newly introduced therapies or updated TNM staging criteria without requiring expensive retraining. The integration of retrieval from trusted clinical sources highlights its potential in dealing with incomplete or ambiguous clinical data.

Graph-based knowledge integration improved the model's reasoning by structuring relationships between clinical entities. Rather than merely retrieving relevant concepts, the knowledge graph enabled the model to place these concepts into a structured context, improving logical reasoning and reducing errors due to ambiguous terms. This structured reasoning aligns closely with clinical workflows, where decisions depend on clear relationships between diagnoses, treatments, and outcomes. By linking predictions to specific nodes in the graph, the model can help with traceability and explainability, which are crucial for building clinician trust.

Model size played a role in performance. Larger models, like \texttt{LLaMA-3.1-8B}, excelled in extracting biomedical entities. However, smaller models like \texttt{LLaMA-2-7B} achieved comparable results on many tasks, particularly when supported by retrieval and graph integration. This trade-off between performance and computational cost is especially relevant for resource-constrained settings. Smaller models, paired with efficient retrieval and graph-based reasoning, offer a viable pathway for deploying advanced NLP tools in clinics with limited hardware capabilities.

Our experiments showed diminishing returns with high instruction counts. After approximately 200 German instructions, improvements plateaued for simpler tasks, such as ICD-10 coding. Complex tasks, like TNM staging, showed marginal gains up to 400 instructions. This finding shows the importance of tailoring instruction counts to task complexity and resource availability. Future exploration of instruction prioritization or curriculum learning could optimize the cost-benefit balance, ensuring that effort is directed where it yields the most significant gains.

The cross-lingual modeling approach demonstrated real-world applicability. Bilingual instruction tuning, combined with retrieval and knowledge graphs, empowered the model to navigate clinical texts in another language, even with minimal supervision. This adaptability can address challenges faced by rural or underserved regions where linguistic diversity often limits access to advanced clinical technologies. Adding a modest number of domain-specific glossaries or synthetic training examples may further enhance performance on rare or compound medical terminology.

Qualitative analyses showed model limitations. On NER tasks, confusion between biomarkers like EGFR and HER2 highlighted the need for more robust contextual disambiguation. Graph-based reasoning mitigated these issues in part by linking terms to authoritative definitions, yet uncommon or rare entities continued to pose challenges. Similarly, for TNM staging extraction, the model excelled with standard terminology but struggled with vague or non-standard formulations. Retrieval partially addressed these gaps by surfacing canonical TNM definitions, while graph integration provided structured connections between terms and staging guidelines. However, cases where clinical texts themselves lacked clarity remained problematic, underscoring the dependence of NLP systems on the quality of source data.

Cross-lingual coding introduced unique challenges. While minimal German instructions helped the model perform ICD-10 coding and SNOMED classification tasks, the model occasionally failed with long compound German words or uncommon clinical expressions. Refining multilingual embeddings and incorporating domain-specific lexicons could improve outcomes, particularly in handling rare terminology and idiomatic language usage.

A deeper look at the model's performance on the \textit{MSK-IMPACT} dataset revealed its ability to correctly match common mutations, such as EGFR, to appropriate therapies. However, the model struggled with rare genetic variants due to sparse retrieval references. In such cases, indirect reasoning and inference from related mutations proved insufficient. Future work could address this limitation by integrating curated genomic knowledge bases or using generative retrieval strategies to synthesize knowledge from related contexts.

Future directions could refine these methods further. Expanding multimodal capabilities by integrating text-based NLP with imaging data, such as radiology scans or histopathology images, could create a more comprehensive oncology assistant. Generative retrieval strategies and graph embedding techniques may raise the performance ceiling by improving the depth and scope of retrieved knowledge. Extending cross-lingual integration to low-resource languages could address global disparities in healthcare technology access. Testing the framework in clinical trials with real-world practitioners will provide critical insights into its usability, reliability, and impact on decision-making.

\bibliography{anthology,custom}
\bibliographystyle{acl_natbib}

\appendix

\end{document}